\documentclass[10pt,twocolumn,letterpaper,dvipsnames]{article}

\usepackage[pagenumbers]{cvpr} 
\usepackage{times}
\usepackage{easyReview}
\usepackage{epsfig}
\usepackage{graphicx}
\usepackage{mathtools}
\usepackage{amssymb}
\usepackage{tikz}
\usepackage{bbding} 
\usepackage[accsupp]{axessibility}

\usepackage{multirow, makecell}
\usepackage{pifont}
\usepackage{xcolor}
\usepackage{pgfplots, adjustbox}
\pgfplotsset{width=10cm,compat=1.9}

\usepgfplotslibrary{external}
\usepackage[]{xcolor}

\newcommand{\Lagr}{\mathcal{L}}

\newcommand{\errmath}[1]{$\scriptscriptstyle \pm #1$}
\newcommand{\yiranresponse}[2][1=]

\definecolor{cvprblue}{rgb}{0.21,0.49,0.74}
\usepackage[pagebackref,breaklinks,colorlinks,citecolor=cvprblue]{hyperref}

\title{Grounding Stylistic Domain Generalization with Quantitative Domain Shift Measures and Synthetic Scene Images}
\author{
Yiran Luo$^{1}$\thanks{Corresponding author} \quad Joshua Feinglass$^{1}$ \quad Tejas Gokhale$^{2}$ \quad Kuan-Cheng Lee$^{1}$ \quad Chitta Baral$^{1}$ \quad Yezhou Yang$^{1}$ \\
$^{1}$Arizona State University \;  $^{2}$University of Maryland, Baltimore County \\
$^{1}${\tt\small \{yluo97, jfeingla, klee201, chitta, yz.yang\}@asu.edu} \; $^{2}${\tt\small gokhale@umbc.edu} \\
}

\usepackage{amsmath,amsfonts,bm}

\def\eqref#1{equation~\ref{#1}}

\def\1{\bm{1}}

\DeclareMathAlphabet{\mathsfit}{\encodingdefault}{\sfdefault}{m}{sl}
\SetMathAlphabet{\mathsfit}{bold}{\encodingdefault}{\sfdefault}{bx}{n}

\usepackage{url}
\usepackage{booktabs, multirow, wrapfig} 
\usepackage{soul}
\usepackage{changepage,threeparttable} 
\usepackage{enumitem}

\def\paperID{14} 
\def\confName{VDU}
\def\confYear{2024}

\begin{document}

\maketitle
\begin{abstract}
Domain Generalization (DG) is a challenging task in machine learning that requires a coherent ability to comprehend shifts across various domains through extraction of domain-invariant features. 
DG performance is typically evaluated by performing image classification in domains of various image styles. 
However, current methodology lacks quantitative understanding about shifts in stylistic domain, and relies on a vast amount of pre-training data, such as ImageNet1K, which are predominantly in photo-realistic style with weakly supervised class labels. 
Such a data-driven practice could potentially result in spurious correlation and inflated performance on DG benchmarks.
In this paper, we introduce a new 3-part DG paradigm to address these risks. 
We first introduce two new quantitative measures \textbf{ICV} and \textbf{IDD} to describe domain shifts in terms of consistency of classes within one domain and similarity between two stylistic domains. 
We then present \textbf{SuperMarioDomains (SMD)}, a novel synthetic multi-domain dataset sampled from video game scenes with more consistent classes and sufficient dissimilarity compared to ImageNet1K. 
We demonstrate our DG method \textbf{SMOS}. SMOS uses SMD to first train a precursor model, which is then used to ground the training on a DG benchmark. 
We observe that SMOS+SMD altogether contributes to state-of-the-art performance across five DG benchmarks, gaining large improvements to performances on abstract domains along with on-par or slight improvements to those on photo-realistic domains.
Our qualitative analysis suggests that these improvements can be attributed to reduced distributional divergence between originally distant domains. 
Our data are available at \url{https://github.com/fpsluozi/SMD-SMOS} .
\vspace{-0.5cm}

\end{abstract}

\section{Introduction}
\begin{figure}[!t]
\vspace{-.2cm}
\includegraphics[width=0.47\textwidth]{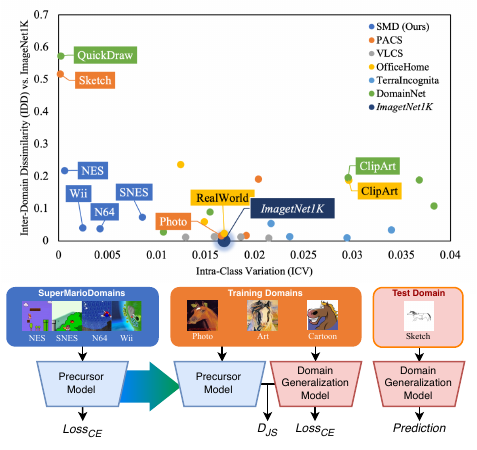} 
\caption{ \textbf{Top: }We define two quantitative measures \textbf{ICV} and \textbf{IDD} to describe stylistic domain shifts in image datasets for Domain Generalization (DG). We find that the vast ImageNet1K, commonly used for pre-training DG models, has inconsistent class labels and is already similar in style with photo-realistic domains found in multiple benchmarks. Therefore, we compile a novel synthetic dataset \textbf{SuperMarioDomains (SMD)} as referential stylistic domains with consistent scene class labels and sufficient dissimilarity from existing domains. 
\textbf{Bottom: } We present our DG approach \textbf{SMOS} that leverages the unique domain shifts in our new \textbf{SMD} dataset. We first train a Precursor Model using SMD and cross entropy $Loss_\mathrm{CE}$. We then utilize the trained Precursor Model to ground the training of the DG model with training domains from the benchmark, optimizing the empirical loss of both cross entropy $Loss_\mathrm{CE}$ and Jensen-Shannon Divergence $D_\mathrm{JS}$ between the Precursor Model and the DG Model.\label{figure1}}



\end{figure}

The generalizability of deep neural networks in computer vision is a crucial task that is still challenging \cite{texture, texture2}.
Hence, the specific task of Domain Generalization (DG) has been defined with the aim of improving the generalization of models by singling out distribution shifts among data that belong to independent and identically distributed (i.i.d) domains \cite{domain_shift, domainbed}. The evaluation of a DG model consists of performing supervised image classification in a multi-source leave-one-out scenario, where one domain is held out as an unseen test set, and other domains are for training. 

The crucial DG strategy is to learn domain-invariant features from the training domains such as SagNet \cite{sagnet}, CORAL \cite{coral2016}, and DANN \cite{dann}. More DG benchmarks of more perplexing stylistic domains and more fine-grained classes \cite{officehome, terraincongnita, wilds, camelyon17, domainnet} have also been developed for more comprehensive evaluation of generalizability. 
However, we notice potential risks in the current methodology. Most DG methods solely rely on initializing their backbone models pre-trained with vast weakly supervised image collections, e.g. ImageNet1K \cite{imagenet}, which overwhelmingly resemble \cite{simple, ermplus} photo-styled domains in multiple DG benchmarks. We observe that many DG methods \cite{erm, irm, coral2016, sagnet, miro} reduce various forms of qualitative distances among domains, but lack quantitative understanding of the specific differences amongst the domains of image styles. Without clarified and unbiased understanding of the common ground among training domain examples, generalization onto an unseen domain would potentially be based upon un-grounded spurious evidence, resulting in inflated DG performance. 



In this paper, we present a new 3-fold paradigm for Domain Generalization. First, we define 2 quantitative measures for stylistic domain shifts based on Jensen-Shannon Divergence \cite{jsd} - \textbf{Intra-Class Variation (ICV)} within an individual domain, and \textbf{Inter-Domain Dissimilarity (IDD)} between two domains. With these 2 measures, as shown in Figure \ref{figure1} in the top, we confirm that ImageNet1K is biased toward domains of photorealistic styles, but also has inconsistent representations within its individual class categories. 


We then construct a new multidomain dataset dubbed \textbf{ SuperMarioDomains (SMD)}
\footnote{As of Mar. 20, 2024, the official content guidelines of Nintendo, who owns the copyright of all games involved in SMD, has clearly stated they 'will not object to your use of gameplay footage and/or screenshots captured from games for which Nintendo owns the copyright ... for appropriate video and image sharing sites' \cite{nintendo}. We will follow this guideline and publish our work in two forms. Extracted feature vectors and pre-trained models will be readily available under the MIT license. The SMD dataset with its raw images will be accessible after agreeing to using the dataset for fair-use research purposes only under the Open Database License (ODbL).  }. 
SMD features 4 domains representing multiple generations of video game graphic styles, ranging from low-resolution pixels to 3D-rendered polygon-rich graphics. All domains share 4 classes of in-game type of scenes that appear consistently throughout the Mario franchise. Unlike ImageNet1K, SMD's domains maintain variable stylistic distances from ImageNet1K in terms of IDD, while having more consistently labeled examples than ImageNet1K's in terms of ICV. 

Our proposed DG method \textbf{SMOS} is shown in the bottom of Figure \ref{figure1}. SMOS first trains a precursor model with domains in SMD. It then trains a DG model grounded by the distribution represented by the SMD-trained precursor model. We apply SMOS and find overall improvements on multiple DG benchmarks. SMOS contributes to the most significant improvements when targeting abstract-styled domains, for example $+7.3\%$ on Sketch of PACS \cite{pacs}, $+3.6\%$ on Clipart of OfficeHome \cite{officehome}, or $+8.1\%$ on Clipart of DomainNet \cite{officehome} over the baselines of MIRO \cite{miro}. SMOS also maintains performance on par with baselines when targeting photo-realistic domains, and in some cases, e.g. PACS, even has improvements.
We also observe that SMOS is able to improve extraction of domain-invariant features and generalization over domain shifts in quality, as we find that originally distant stylistic domains are now projected within considerably smaller distributional divergence. 

Our main contributions in this paper are as follows.
\begin{enumerate}
    \item  We propose ICV and IDD as measures for stylistic domain shifts in DG benchmarks. Using our measures, we find that real-world datasets like ImageNet1K used as pre-training data for DG may not be ideal for obtaining domain-invariant features due to inconsistent classes and overt stylistic similarity with training domains. 
    \item We introduce a new synthetic dataset SuperMarioDomains as a precursor dataset for DG, incorporating unique features of consistent classes of video game scenes across stylistic domains in video game graphics that are dissimilar to ImageNet1K. 
    \item We present our DG method SMOS. SMOS utilizes SuperMarioDomains to obtain a precursor model that grounds the training process on DG benchmarks. We show that SMOS is capable of obtaining top performance on multiple DG benchmarks, in particular, via large improvements on targeting abstract-styled domains together with on-par or slight improvements on photorealistic-styled domains. 
\end{enumerate}

\section{Related Works}
\begin{table}[!t]
\centering
\small
\begin{tabular}{lllr}\toprule
Dataset & Domains & Classes &  Size  \\\midrule
ImageNet1K \cite{imagenet}& 1 image style & 1K objects & 1.3M \\
Places365 \cite{places}& 1 image style& 365 scenes & 1.8M \\
\midrule
PACS \cite{pacs}& 4 image styles & 7 objects &10K   \\
VLCS \cite{vlcs}& 4 photo sources & 5 objects &10K    \\
TerraIncognita \cite{terraincongnita} & 4 locations & 10 animals & 25K \\
OfficeHome \cite{officehome}&4 image styles &65 objects &16K   \\
DomainNet \cite{domainnet}&6 image styles &345 objects & 587K  \\
\midrule
SMD (ours)& 4 game graphics & 4 scenes& 82K \\
\bottomrule
\end{tabular}
\newline
\caption{Statistics of image classification datasets involved in DG. }\label{tab: stats_dataset}
\vspace*{-1\baselineskip}
\end{table}

\noindent\textbf{Synthetic Datasets with Domain Shifts.}
Synthetic data has long been used in various disciplines in computer vision \cite{syntheticdatasets,metasimulation,ovvv,gym, achddou2021synthetic, courtois2022investigating}. Recently, we have seen that more synthetic datasets of distribution shifting domains are being assembled to encourage more robust algorithms in different tasks. We have benchmarks studying adaptation between synthesized and real-life objects in Syn2Real \cite{syn2real} or SYNTHIA \cite{ros2016synthia}. Super-CLEVR \cite{superclevr} studies more robust visual reasoning skills by constructing domains w.r.t. visual question answering settings. More specifically, we draw great inspiration from the GTAV-Cityscapes challenge \cite{gta, cityscapes}, adapting from a vast collection of video game landscapes to real-world scenarios for segmentation. All these works show that synthetic datasets, though lacking full realism, may help provide great insight into domain shifts.

\noindent\textbf{Domain Generalization Benchmarks.}
Early DG benchmarks such as Office \cite{office} or VLCS \cite{vlcs} focus solely on photo-realistic images. Since the time when single-style bias was first exposed by DeCAF \cite{decaf}, more fine-grained image styles have been introduced to the mix of image domains, and we have seen a steady increase in scale for DG datasets, including OfficeHome \cite{officehome}, PACS \cite{pacs}, TerraIncognita \cite{terraincongnita}, SVIRO \cite{sviro}, WILDS \cite{wilds}, Camelyon17 \cite{camelyon17}, and NICO++ \cite{nico++}. Currently, the DomainNet dataset \cite{domainnet} is the largest with 587K images, featuring 6 stylistic domains and 345 object categories.

\noindent\textbf{Domain Generalization Approaches.}
Techniques to tackle the problem of domain shift in Domain Generalization have been rapidly developed over the years. Researchers are no longer restricted to straightforward approaches such as finding linear \cite{linearalign} representations with techniques like interpolation \cite{mixup,cutmix2019,covariate2022,smurf2021} or nonlinear \cite{nonlinearalign} representations in-between domains. In many circumstances, simple methods, such as variants of empirical risk minimization ~\cite{gulrajani2021in, irm}, can produce high performance on popular domain generalization benchmarks. But recent approaches to domain generalization can involve adaptation and combination of deep neural network models such as DANN \cite{dann} and CDANN \cite{cdann}, leveraging Meta-Learning \cite{metalearning} or Adversarial-Learning \cite{volpi2018generalizing,qiao2020learning,alt} to transfer model parameters, using an ensemble of different architectures \cite{simple}, training strategies \cite{jiang2023domain}, or regularization methods to improve generalizability like SWAD \cite{swad} or MIRO \cite{miro}.

\label{chapter: preliminaries}
\begin{figure*}[t]
    \centering
    \includegraphics[width=0.85\linewidth]{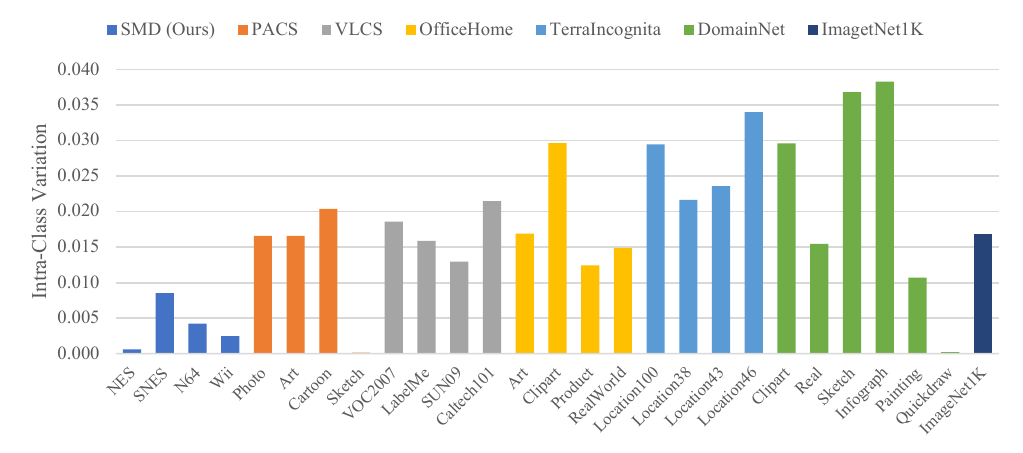}
    \caption{\textbf{Intra-Class Variation (ICV)} for each domain in featured datasets. A low ICV indicates that the classes are more consistent in terms of colors, as in NES of SMD, Sketch of PACS, and Quickdraw of DomainNet. Meanwhile, the classes in ImageNet1K, which is commonly used for pre-training in DG, are implicated to be as inconsistent as those in photo- or art-styled domains, e.g. Photo and Art of PACS, LabelMe of VLCS, Art and RealWorld of OfficeHome, as well as Real of DomainNet. Average of 3 trials.}
    
    \label{fig:domain-intra}
    \includegraphics[width=0.85\linewidth]{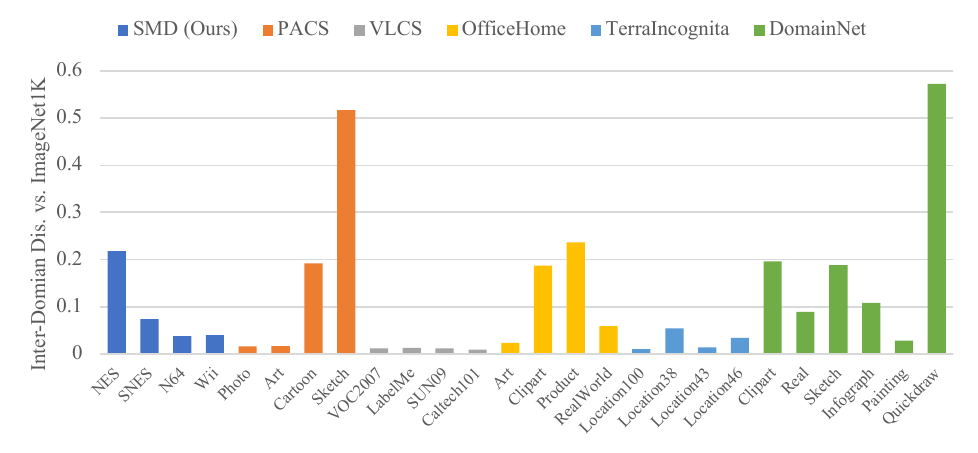}
    \caption{\textbf{Inter-Domain Dissimilarity (IDD)} of ImageNet1K vs. each domain in featured datasets. IDD of ImageNet1K vs. itself is presumably 0. Since ImageNet1K is dominantly real-world photographs, those with smaller IDD implicate stronger resemblance to the style of photo-realism, e.g. all 4 domains of VLCS, or Photo of PACS. Highly abstract and simplistic styles, such as Sketch of PACS and Quickdraw of DomainNet, are shown in very large IDD values on the flip side. }
    \label{fig:domain-dis}
\end{figure*}

\section{Preliminaries}

In this paper, we focus on the Domain Generalization (DG) performance of image classification by conducting Multi-Source Domain Generalization (MSDG), where we evaluate a model's performance with \textit{leave-one-out cross-validation}. In formal terms, a DG dataset $\mathcal{D}$ is divided into $N$ domains $\{d_1, ... d_{N}\}$. Each individual domain $d_i$ contains images $X^{d_i}$ paired with their class labels $Y^{d_i}$. All domains share the same set of image labels. We train a model $\mathcal{M}$ with the training set (training domains) $\mathcal{D}_{tr}$ that uses all data from all-but-one domains $\mathcal{D}_{tr} = \{(X^{d_1}, Y^{d_1}), ...  (X^{d_{N-1}}, Y^{d_{N-1}})\}$. The trained model is then evaluated using unseen data from the one held-out test (target) domain as the test set $\mathcal{D}_{te} = \{(X^{d_{N}}, Y^{d_{N}})\}$. Noticeably, unlike Domain Adaptation \cite{da}, the one test domain $\mathcal{D}_{te}$ in DG is not involved in any training and does not contain samples that overlap with the training domains. 

A typical DG model $\mathcal{M} = f \circ g$ consists of two core components: a feature extractor (featurizer) $f$ as the backbone network, followed by a single-layered linear classifier $g$. Although more complicated methods have been developed over the years, the fundamental approach of \textbf{Empirical Risk Minimization} (\textbf{ERM}) \cite{erm} remains relevant. In the context of DG, the goal of ERM is simply minimizing the classification loss averaged over the $N-1$ training domains given training data $x_i, y_i\in \mathcal{D}_{tr}$ :
\begin{equation}
\Lagr_{\mathrm{ERM}}= \frac{1}{N-1} \sum_{x_i, y_i\in \mathcal{D}_{tr}}\Lagr_{\mathrm{CE}}(f \circ g (x_i), y_i).
\label{eq: erm}
\end{equation}
Although straightforward, the latest baseline ERM \cite{gulrajani2021in}, which applies cross-entropy as the training loss $\Lagr_{\mathrm{CE}}$, is shown to rank high in many DG benchmarks \cite{ermplus, irm2022, domainbed}.

\label{chapter: domain_distances}

\section{Analyses on Domain Shift of Stylistic Domains and Pre-training Data}
\label{chapter: js-distance}
\textbf{Risk in Pre-training Data Used for DG.} Most DG methods do not train their model $\mathcal{M}$ entirely from scratch. The model's initial weights are commonly transferred from an existing model, typically pre-trained on ImageNet1K \cite{imagenet}. Researchers \cite{ermplus, nico++} have observed that there is an apparent stylistic resemblance between ImageNet1K and many domains from various DG benchmarks, such as the Photo domain in PACS\cite{pacs}, the RealWorld domain in OfficeHome\cite{officehome}, or even all 4 domains in VLCS \cite{vlcs}. 

Such a common practice of only depending on a vast real-world image dataset can be risky. Qualitatively, the images used for pre-training lack shift in style to adapt from, e.g. ImageNet1K are dominated by real-world photos. In addition, the images in ImageNet1K may be inconsistent within the same class; e.g., images of a class of fish may or may not include a person holding it. Hence, when we perform on DG benchmarks by using a model pre-trained only by such data, we may find it difficult to learn coherent style-invariant presentations in order to generalize over distributional shifts among domains, especially when we have to train with the predominating photo-realistic domains and test on an unseen abstract-styled domain.

\begin{figure*}[!t]
\centering
\includegraphics[width=0.80\textwidth]{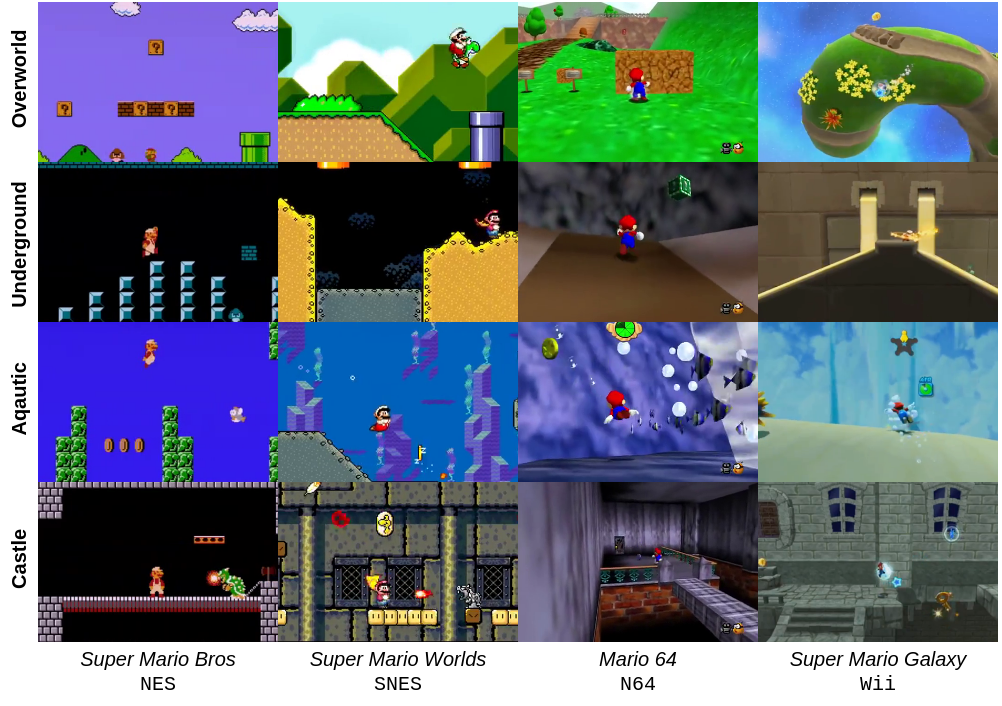} 
\vspace{-.2cm}
\caption{A qualitative overview of our \textbf{SuperMarioDomains(SMD) }dataset, consisting of video frames from actual game footage categorized into 4 distinctive scene classes and 4 image style domains. \textbf{Columns from left to right:} The four image domains, named after the console hardware on which each game runs - $\mathtt{NES} $, $\mathtt{SNES} $, $\mathtt{N64} $, and $\mathtt{Wii} $. \textbf{Rows from top to bottom:} The four classes of in-game scenes - Overworld, Underground, Aquatic, and Castle. These synthetic image styles of SMD are drastically different from those in existing DG benchmarks, such as realistic photographs, pencil sketches, or oil paintings.
}\label{overview}
\end{figure*}

\noindent\textbf{Quantitative Measures of Stylistic Domain Shift.} We would next like to compare the advantages of different DG pre-training data like ImageNet1K. Inspired by previous works which examine image similarity \cite{image-similarity, jsnet}, we define two intuitive measures for domain shift in DG: \textbf{Intra-Class Variation (ICV)} to implicate the class-wise presentation inconsistency within one specific domain, and \textbf{Inter-Domain Dissimilarity (IDD)} to implicate the distance in representation distributions across two domains of styles. 

Both measures utilize the symmetric Jensen-Shannon Divergence (JSD) \cite{jsd}. Concretely, let \(P\) and \(Q\) be the estimated probability distributions of the RGB channels and bins (3$\times$256 dimensions), making the outcome set a vector of length 768. The JSD between \(P\) and \(Q\) is defined as:
\begin{equation}
    D_{\mathrm{JS}}(P, Q) = \frac{1}{2} \left(D_{\mathrm{KL}}(P||M) + D_{\mathrm{KL}}(Q||M) \right),
\label{eq: js}
\end{equation}
where \(M\) is the average mixed distribution:
\begin{equation}
M = \frac{1}{2} (P + Q),
\end{equation}
and the KL divergence between two distributions \(P\) and \(Q\):
\begin{equation}
D_{\mathrm{KL}}(P||Q) = \sum_{x} P(x) \log \frac{P(x)}{Q(x)},
\label{eq: kl}
\end{equation}
where \(P(x)\) and \(Q(x)\) are the probabilities of image instance \(x\) in distributions \(P\) and \(Q\) respectively. $D_{\mathrm{JS}}$ ranges between 0 and 1, with 0 indicating identical distributions and 1 indicating completely dissimilar distributions.

For \textbf{Intra-Class Variation} of a given image domain, \(P_i\) and \(Q_i\) represent distributions of two equal splits of samples belonging to the same class $i$. The \textbf{ICV} w.r.t. the domain $\mathcal{D}$ of $n$ classes is the average over all its class-wise JSDs:

\begin{equation}
     ICV(\mathcal{D}) = \frac{1}{n}\sum_{i=1}^{n} D_{\mathrm{JS}}(P_i, Q_i).
\label{eq: icv}
\end{equation}

For \textbf{Inter-Domain Dissimilarity} between two domains $\mathcal{D}_1$ and $\mathcal{D}_2$, the metric straightforwardly computes the JSD between respective distributions \(P_{\mathcal{D}_1}\) and \(P_{\mathcal{D}_2}\):

\begin{equation}
     IDD(\mathcal{D}_1, \mathcal{D}_2) = D_{\mathrm{JS}}(P_{\mathcal{D}_1}, P_{\mathcal{D}_2}).
\label{eq: icv}
\end{equation}

For fairness, we assume that all image presentations follow Gaussian distributions with regard to probabilities of raw RGB values within bins of $[0, 255]$, normalized by the uniform mean $[0.5, 0.5, 0.5]$ and standard deviation $[0.5, 0.5, 0.5]$. In addition to ImageNet1K, we choose to study the domains in the following DG benchmarks: PACS, VLCS, OfficeHome, TerraIncognita, and DomainNet. The ICV for each domain studied is averaged over 3 trials.

In Figure \ref{fig:domain-intra}, we first compare the ICV of ImageNet1K with that of every domain in the chosen DG benchmarks. We discover that, in ImagetNet1K, image representations within the same class can be as diversely distributed as those learning samples from photo- or art-styled domains in DG benchmarks. Meanwhile, domains of highly abstract styles, such as Sketch in PACS and Quickdraw in DomainNet, have drastically low ICV values compared to other domains in their respective benchmarks. According to Table \ref{tab: stats_dataset}, since ImageNet1K also has at least twice as many classes compared to DG domains, generalization with known samples solely from high-ICV classes would be difficult to learn domain-invariant features, especially when the test domain samples have vastly low variation in feature space.

From Figure \ref{fig:domain-dis}, we now have a clear view of ImageNet1K's similarity to existing styles in multiple DG benchmarks. This can be observed, for example, from the next-to-0 IDD values of all four domains of VLCS, the Photo and Art domains of PACS, or the Location38 domain of TerraIncognita. In contrast, styles of abstract color patterns or textures, such as the Sketch domain of PACS and the Quickdraw domain of DomainNet, have large IDD values which suggests large distances from ImageNet1K.


\section{Methodology}
\label{chapter: miro2}


\noindent \textbf{The SuperMarioDomains Dataset.} Taking inspirations from previous efforts to construct synthetic datasets with environments rendered in video games\cite{gym, ovvv, gta}, we compile a multi-domain dataset \textbf{SuperMarioDomains (SMD)} from video game scenes that can be leveraged as precursor data for Domain Generalization experiments. As its name suggests, our SMD dataset feature synthetic in-game scenes captured from multiple games in the Mario franchise, with image style domains encompass pixelated mosaic 2D graphics in only 50 colors, all the way to high-polygon 3D graphics in 32-bit colors. We land on 4 classes of distinct in-game scenes that consistently appear in all Mario games of our choice. Each scene class has its own unique traits defined by the combination of terrains, objects, and texture. To ensure high consistency in image styles and class labels, we sample our images from video frames of actual game play footage, - neighboring frames of the same game segment share the same scene class. An overview of the samples in SMD is available in Figure \ref{overview}. 

Quantitatively, SMD is designed to counterbalance the stylistic biases of ImageNet1K by having domains that are \textbf{low in ICV} and relatively \textbf{high in IDD}. On the left of Figure \ref{fig:domain-intra}, we find that the 4 synthetic-styled domains in SMD have much higher class consistency than ImageNet1K, as in generally lower ICVs. Also, in Figure \ref{fig:domain-dis}, SMD's domains also keep a series of evenly placed domain shifts in terms of IDD from the dominating style of ImageNet1K, with NES being the most distant domain while N64 and Wii not as close as Photo of PACS or Caltech101 of VLCS.

\begin{figure}[!t]
\centering
\includegraphics[width=0.47\textwidth]{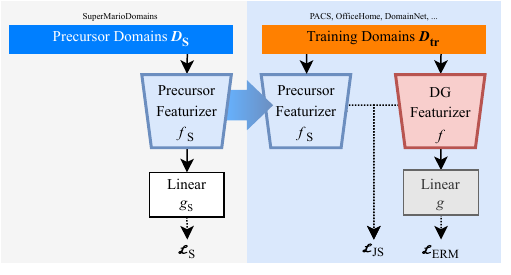} 
\vspace{-.3cm}
\caption{The pipeline of our \textbf{SMOS} method. The feature extraction backbones $f$ and $f_{\mathrm{S}}$ have an identical structure. $f$ is initialized with ImageNet1K pre-trained weights. 
\textbf{Left:} We first train the Precursor Model $f_{\mathrm{S}}\circ g_{\mathrm{S}}$ to learn scene style shifts with SMD. \textbf{Right: }We then perform DG training with training domains from a DG benchmark (e.g. PACS), tuning the DG model $f \circ g$ while being grounded to the SMD-trained Precursor $f_{\mathrm{S}}$ by optimizing $\Lagr_{\mathrm{JS}}$. }
\vspace{-.0cm}
\label{SMIRO}
\end{figure} 

\noindent\textbf{The SMOS DG Method.} Seeing the unique stylistic domain shifts in SMD, we are motivated to design a new DG method with better domain-invariant feature extraction, learning the domain shifts first from isolated class-consistent samples in SMD. We present the \textbf{S}cene-grounded \textbf{M}inimal d\textbf{O}main \textbf{S}hift (\textbf{SMOS}) method to best leverage the unique domain shifts in SMD. Figure \ref{SMIRO} demonstrates the SMOS pipeline. We first train a Precursor Model $\mathcal{M}_{\mathrm{S}} = f_{\mathrm{S}} \circ g_{\mathrm{S}}$ with a Precursor Feature Extractor $f_{\mathrm{S}}$ to learn about the domain shifts of SMD. Sequentially, we train an identically structured model $\mathcal{M} = f \circ g$ with training domains from a DG benchmark, while simultaneously grounded by minimizing the Jensen-Shannon Divergence between distributions of the corresponding feature extractors $f_{\mathrm{S}}$ and $f$.

Formally, the distribution of the Precursor Model $\mathcal{M}_{\mathrm{S}} $ is learned with the synthetic scene data $\mathcal{D}_{\mathrm{S}}$, namely from SMD. Given $N_\mathrm{S}$ stylistic domains of scene image-label pairs $x_j, y_j \in \mathcal{D}_{\mathrm{S} }$, we optimize the cross-entropy loss for updating $\mathcal{M}_{\mathrm{S}}$ in a similar form with ERM:

\begin{equation}
\Lagr_{\mathrm{S}}  = \frac{1}{N_\mathrm{S}} \sum_{x_j, y_j\in \mathcal{D}_{\mathrm{S}}}\Lagr_{\mathrm{CE}}(f_{\mathrm{S}}\circ g_{\mathrm{S}} (x_j), y_j).
\label{eq: bg}
\end{equation}

Together with $N$ training domain images $x_i \in \mathcal{D}_{\mathrm{tr}}$, SMOS additionally grounds the training of the DG model $\mathcal{M}$ by minimizing the JSD from the corresponding Precursor Feature Extractor $f_{\mathrm{S}}$:
\begin{equation}
\Lagr_{\mathrm{JS}} = D_{\mathrm{JS}}(f_{\mathrm{S}}(x_i), f(x_i)).
\label{eq: ground}
\end{equation}
SMOS optimizes the empirical loss during benchmarking:
\begin{equation}
\label{eq: smiro}
\Lagr_{\mathrm{SMOS}} = \Lagr_{\mathrm{S}} + \Lagr_{\mathrm{ERM}} 
+ \lambda \Lagr_{\mathrm{JS}},
\end{equation}
where the coefficient $\lambda$ is a hyper-parameter that controls the grounding factor.

\section{Experiments and Results}

\begin{table*}[!t]
\centering

\scriptsize

\begin{tabular}{l|rrrr|rrr|rrrrr}\toprule
\multirow{3}{*}{\textbf{Method}} &\multicolumn{4}{c|}{PACS} &\multicolumn{3}{c|}{OfficeHome} &\multicolumn{4}{c}{DomainNet} \\\cmidrule{2-12}
&ACS $\rightarrow$ &PAS $\rightarrow$ &PAC $\rightarrow$ &\multirow{2}{*}{Avg.} &ACP $\rightarrow$ &RAP $\rightarrow$ &\multirow{2}{*}{Avg.} &CSIPQ $\rightarrow$ &RSIPQ $\rightarrow$ &CRSIP $\rightarrow$ &\multirow{2}{*}{Avg.} \\
&Photo &Cartoon &Sketch & &RealWorld &Clipart & &Real &Clipart &Quickdraw & \\\midrule
$\mathrm{ERM}$ \cite{erm} &97.0\errmath{0.1}&79.8\errmath{1.0}&74.7\errmath{0.7} &84.2 &77.4\errmath{0.2} &53.2\errmath{1.2} &67.6 &60.9\errmath{0.1} &50.8\errmath{0.2} &10.5\errmath{0.1} &44.0 \\
$\mathrm{MIRO}$ \cite{miro} &97.3\errmath{0.2} &80.5\errmath{0.5} &75.9\errmath{1.0} &85.4 &\textbf{81.1}\errmath{0.4} &53.8\errmath{0.8} &70.5 &63.4\errmath{0.2} &54.2\errmath{0.8} &11.3\errmath{0.0} &44.3 \\
$\mathrm{SMOS}^{-}$(ours) &\textbf{98.1}\errmath{0.0} &\textbf{84.8}\errmath{0.2} &81.3\errmath{0.4} &88.7 &80.7\errmath{0.3} &57.5\errmath{0.3} &71.0 &63.9\errmath{0.3} &60.3\errmath{0.1} &13.9\errmath{0.0} &44.5 \\
$\mathrm{SMOS}^{+}$(ours) &\textbf{98.1}\errmath{0.1} &\textbf{84.8}\errmath{0.3} &\textbf{83.2}\errmath{0.6} &\textbf{89.4} &80.8\errmath{0.4} &\textbf{58.6}\errmath{0.4} &\textbf{71.6} &\textbf{64.0}\errmath{0.2} &\textbf{62.3}\errmath{0.0} &\textbf{14.7}\errmath{0.0} &\textbf{45.3} \\
\bottomrule
\end{tabular}

\caption{DG performance on individual target domains. SMOS gains the greatest improvements on abstract-styled domains while maintaining its performance on photo-styled domains. The letters that precede $\rightarrow$ denote initial letters of the training domains in the respective benchmarks. The \textbf{best results} per setting are shown in bold. Average of 3 trials. }
\label{tab: SMIRO-results-details}
 
\vspace*{-0\baselineskip}
\end{table*}

\begin{table*}[!t]
\centering
\small
\begin{tabular}{l|ccccc|r}\toprule
\textbf{Method} &  PACS & VLCS & OfficeHome & TerraIncognita & DomainNet & Avg. \\\midrule
$\mathrm{Mixstyle}^{\dagger}$~\cite{mixstyle2021} & 85.2 \errmath{0.3} & 77.9 \errmath{0.5} & 60.4 \errmath{0.3} & 44.0 \errmath{0.7} & 34.0 \errmath{0.1} & 60.3 \\
$\mathrm{GroupDRO}^{\dagger}$~\cite{groupdro2020}  & 84.4 \errmath{0.8} & 76.7 \errmath{0.6} & 66.0 \errmath{0.7} & 43.2 \errmath{1.1} & 33.3 \errmath{0.2} & 60.7 \\
$\mathrm{IRM}^{\dagger}$~\cite{irm2022}      & 83.5 \errmath{0.8} & 78.5 \errmath{0.5} & 64.3 \errmath{2.2} & 47.6 \errmath{0.8} & 33.9 \errmath{2.8} & 61.6 \\
$\mathrm{ARM}^{\dagger}$~\cite{arm2021} & 85.1 \errmath{0.4} & 77.6 \errmath{0.3} & 64.8 \errmath{0.3} & 45.5 \errmath{0.3} & 35.5 \errmath{0.2} & 61.7 \\
$\mathrm{VREx}^{\dagger}$ ~\cite{VREx}     & 84.9 \errmath{0.6} & 78.3 \errmath{0.2} & 66.4 \errmath{0.6} & 46.4 \errmath{0.6} & 33.6 \errmath{2.9} & 61.9 \\
$\mathrm{CDANN}^{\dagger}$ ~\cite{cdann}    & 82.6 \errmath{0.9} & 77.5 \errmath{0.1} & 65.8 \errmath{1.3} & 45.8 \errmath{1.6} & 38.3 \errmath{0.3} & 62.0 \\
$\mathrm{DANN}^{\dagger}$ ~\cite{dann}     & 83.6 \errmath{0.4} & 78.6 \errmath{0.4} & 65.9 \errmath{0.6} & 46.7 \errmath{0.5} & 38.3 \errmath{0.1} & 62.6 \\
$\mathrm{RSC}^{\dagger}$ ~\cite{rsc}      & 85.2 \errmath{0.9} & 77.1 \errmath{0.5} & 65.5 \errmath{0.9} & 46.6 \errmath{1.0} & 38.9 \errmath{0.5} & 62.7 \\
$\mathrm{MTL}^{\dagger}$~\cite{mtl2021}     & 84.6 \errmath{0.5} & 77.2 \errmath{0.4} & 66.4 \errmath{0.5} & 45.6 \errmath{1.2} & 40.6 \errmath{0.1} & 62.9 \\
$\mathrm{Mixup}^{\dagger}$~\cite{mixup}     & 84.6 \errmath{0.6} & 77.4 \errmath{0.6} & 68.1 \errmath{0.3} & 47.9 \errmath{0.8} & 39.2 \errmath{0.1} & 63.4 \\
$\mathrm{MLDG}^{\dagger}$~\cite{mldg}      & 84.9 \errmath{1.0} & 77.2 \errmath{0.4} & 66.8 \errmath{0.6} & 47.7 \errmath{0.9} & 41.2 \errmath{0.1} & 63.6 \\
$\mathrm{ERM}^{\dagger}$~\cite{erm}       & 84.2 \errmath{0.1} & 77.3 \errmath{0.1} & 67.6 \errmath{0.2} & 47.8 \errmath{0.6} & 44.0 \errmath{0.1} & 64.2 \\
$\mathrm{SagNet}^{\dagger}$~\cite{sagnet2021}& 86.3 \errmath{0.2} & 77.8 \errmath{0.5} & 68.1 \errmath{0.1} & 48.6 \errmath{1.0} & 40.3 \errmath{0.1} & 64.2 \\
$\mathrm{CORAL}^{\dagger}$~\cite{coral2016} & 86.2 \errmath{0.3} & 78.8 \errmath{0.6} & 68.7 \errmath{0.3} & 47.6 \errmath{1.0} & 41.5 \errmath{0.1} & 64.5 \\
$\mathrm{CCFP}$~\cite{jiang2023domain} & 86.6 \errmath{0.2} & 78.9 \errmath{0.3} & 68.9 \errmath{0.1} & 48.6 \errmath{0.4} & 41.7 \errmath{0.0} & 64.8 \\
$\mathrm{MIRO}^{\dagger}$~\cite{miro}& 85.4 \errmath{0.4} & 79.0 \errmath{0.0} & 70.5 \errmath{0.4} & 50.4 \errmath{1.1} & 44.3 \errmath{0.2} & 65.9 \\
\midrule
$\mathrm{SMOS}^{-}$(ours)           &  88.7 \errmath{0.2} &79.7 \errmath{0.1} &71.0 \errmath{0.0} &\textbf{55.5} \errmath{0.8} &44.5 \errmath{0.0} & 67.9\\
$\mathrm{SMOS}^{+}$(ours)            & \textbf{89.4} \errmath{0.3} &\textbf{79.8} \errmath{0.1} &\textbf{71.6} \errmath{0.1} &55.4 \errmath{0.4} &\textbf{45.3} \errmath{0.0} & \textbf{68.3} \\
\bottomrule
\end{tabular}

\caption{Average leave-one-out cross-validation performances on multiple DG benchmarks. $^{\dagger}$are baseline results reported in \cite{miro}. 
The \textbf{best results} per DG benchmark are highlighted in bold. All accuracies and errors are averaged from 3 trials. }\label{tab: SMIRO-results}

\end{table*}

\noindent\textbf{Experiment setups.} We use ResNet50 \cite{resnet} pre-trained with ImageNet1K \cite{imagenet} as the single default backbone for feature extractor networks. Our implementation is based on the source code of DomainBed \cite{gulrajani2021in} and MIRO \cite{miro}. We evaluate the performance of SMOS on the following DG benchmarks: PACS \cite{pacs}, OfficeHome \cite{officehome}, VLCS \cite{vlcs}, TerraIncongnita \cite{terraincongnita}, and DomainNet \cite{domainnet}.

Two variants of SMOS are designed to initialize $\mathcal{M}_{\mathrm{S}}$ differently - \textbf{SMOS$^{-}$} whose $f_{\mathrm{S}}$ is initialized from scratch via Kaiming Initialization \cite{kaiminginit}, and \textbf{SMOS$^{+}$} whose $f_{\mathrm{S}}$ is initialized with ImageNet1K pre-trained weights.

\begin{table}[t]
\centering
\footnotesize
\resizebox{\linewidth}{!}{
\begin{tabular}{lrrrrr} 
 \toprule
 \textbf{Hyperparameter}& PACS & OH & DN & TI & VLCS \\ \midrule
 Learning rate&{3e-5}&{3e-5}&{3e-5}&{3e-5}& 1e-5\\ 
 Dropout & 0.0 & 0.1 & 0.1&0.0 &0.5\\ 
 Weight decay&0.0 & 1e-6& 1e-4 &0.0 & 0.0\\ 
Steps & 5000 & 5000 & 15000 & 5000 & 5000 \\
 $\lambda$ &0.15 & 0.1&0.1 & 0.1 &0.01\\ 
\bottomrule
\end{tabular}
}
\caption{Hyperparameters for DG experiments. OH, DN, and TI respectively stand for OfficeHome, DomainNet, and TerraIncognita. $\lambda$ is our grounding coefficient for SMOS as in Equation \ref{eq: smiro}.}
\label{table: hp1}
\end{table}

We apply the same data augmentations by ERM \cite{gulrajani2021in} and MIRO \cite{miro} - each training set image is cropped with random size and aspect ratio, resized to 224×224 pixels, applied random horizontal flips, applied random color jitter, applied grayscale with 10\% probability, and normalized using the ImageNet channel means and standard deviations. For hyperparameters, we use a batch size of 16, and the Adam optimizer\cite{kingma2014adam} for all experiments. The hyperparameters listed in Table \ref{table: hp1} are consistent with those used in the MIRO paper \cite{miro}, where for each benchmark, we use a different combination of hyperparameters. We divide all domains in SMD by a 4-to-1 training-test ratio to obtain the precursor model $\mathcal{M}_{\mathrm{S}}$ in SMOS. All experiments are conducted using 2 NVIDIA V100 GPUs.

\label{section-SMOS}


\noindent\textbf{Domain Generalization Performance with SMOS.} We present the most significant improvements of SMOS on the individual target domains in Table \ref{tab: SMIRO-results-details}. The key takeaway is that our SMOS method demonstrates its best strength when generalizing onto more abstract-styled domains, e.g. Sketch in PACS, on which baseline methods such as ERM and MIRO struggle to improve. Using the unique domain shifts in SMD, our SMOS$^{+}$ variant achieves large improvements over the state-of-the-art MIRO method by $+3.7\%$ and $+7.3\%$ on Cartoon and Sketch in PACS, $+3.6\%$ on Clipart in OfficeHome, and $+8.1\%$ and $+3.4\%$ on Clipart and Quickdraw in DomainNet. We also show that SMOS is able to maintain its performance on photo-realistic domains on par within the MIRO baseline's error range. In some cases, such as PACS and DomainNet, SMOS can even gain slightly better performance than the MIRO baseline.

A comprehensive comparison of a larger selection of DG methods on multiple benchmarks is shown in Table \ref{tab: SMIRO-results}. We show that our 2 SMOS variants achieve better overall performance on the chosen DG benchmarks, thanks to the particular improvements on abstract-styled domains. Our SMOS$^{+}$ variant surpasses the baseline MIRO by $+4.0\%$ on PACS, $+0.9\%$ on VLCS, $+1.1\%$ on OfficeHome, $+4.9\%$ on TerraIncognita, and $+1.0\%$ on DomainNet.

\noindent\textbf{Qualitative Analysis of Domain-invariant Feature Extraction.} We further validate SMOS' generalizability, showing it helps project originally distant domains to closer distributional proximity, which qualitatively implicates improvements in extracting domain-invariant features and thus better grounded DG performance. We apply our divergence-based IDD measure between domains that belong to the same benchmark, comparing distributions of representations extracted by different DG methods. 

\begin{figure}[t]
    \centering
    \includegraphics[width=\linewidth]{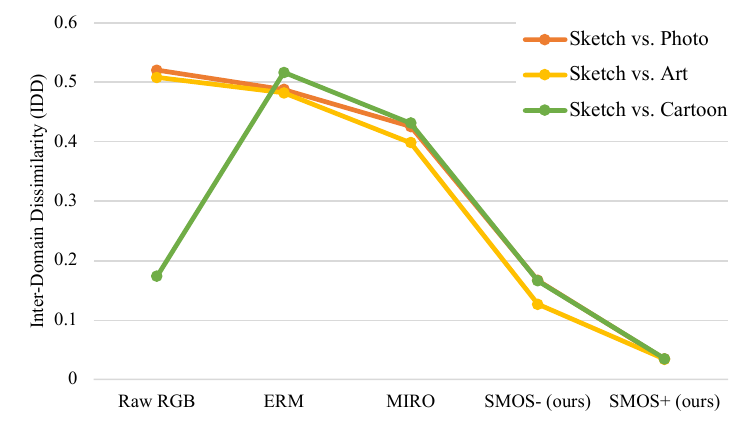}
    \caption{Test vs. training domain IDDs resulted from different DG methods when targeting the Sketch domain of PACS.}
    \label{fig:proximity}
\end{figure}

Figure \ref{fig:proximity} exemplifies the case from the perspective of the Sketch domain in PACS. In raw RGB distributions, Sketch is shown to be highly distant from other domains except Cartoon. When conducting DG image classification tests on Sketch, baseline DG methods such as ERM and MIRO are shown to incrementally lower the distributional divergence from Photo and Art, but at the cost of Cartoon which is originally the most similar to Sketch in RGB. Our methods SMOS$^{-}$ and SMOS$^{+}$, along with our SMD precursor data, universally lower the divergences with all non-Sketch domains of PACS. Our methods are thus able to extract domain-agnostic representations within much closer distributional proximity in respective feature spaces, presented as further improved DG performance on Sketch. 

\begin{table}[t]\centering
\small
\begin{tabular}{l|l|cc}\toprule
\multirow{2}{*}{\textbf{Method}}&\multicolumn{1}{c|}{Precursor}&\multicolumn{2}{c}{Benchmark Avg. }  \\
&\multicolumn{1}{c|}{ Train. Data $\mathcal{D}_{\mathrm{S}}$}  &PACS & OfficeHome\\\midrule

$\mathrm{ERM}$   &-                 &84.2 &67.6\\
$\mathrm{MIRO}$   &-                 &85.4 &70.5\\\midrule   
$\mathrm{SMOS}^{-}$           &DomainNet            & 87.7 & 65.2\\ 
             &Places365     & 87.4 & 66.5\\  
            &SMD (ours)            & 88.7 & 71.0 \\  \midrule
$\mathrm{SMOS}^{+}$        &DomainNet            & 87.9 & 70.1 \\ 
            &Places365     & 87.5 & 66.5 \\  
            &SMD (ours)           &\textbf{89.4} & \textbf{71.6}  \\ 

\bottomrule

\end{tabular}
\caption{DG performance of using substitute precursor data in SMOS$^{-}$ and SMOS$^{+}$. 
All substitute precursor data are downsampled to the same size of SMD (80k). Average of 3 trials.} \label{table: ablation-places}
 \vspace*{-0.7\baselineskip}
\end{table}

\noindent\textbf{Ablation Study of Substitute Precursor Data with SMOS.} We also experiment with using substitute image classification datasets as precursor data ($\mathcal{D}_{\mathrm{S}}$ as in Figure \ref{SMIRO}) under the SMOS method, while SMD is not involved. For substitutes, we choose the benchmark dataset DomainNet \cite{domainnet} which, on its own, has more different domains of image styles and more classes compared to SMD, but also far higher variance in both ICV and IDD than SMD (shown in Figure \ref{figure1}). We also choose the scene-based Places365-Standard (Places365) dataset \cite{zhou2017places} that has 300+ classes of scenes similar to SMD, but only in one photo-realism style identical to ImageNet1K. We randomly downsample the substitute precursor data to the same size of SMD (80k) at each trial. We use the same 4-to-1 training-test split ratio when obtaining alternative precursor models. We perform benchmarking on PACS and OfficeHome. 

Table \ref{table: ablation-places} shows the benchmarking performances of using different precursor data in the SMOS paradigm. We find that SMOS retains the best DG performance when it utilizes SMD's unique stylistic domain shifts and scene labels. In contrast, we see that training the precursor model with ImageNet1K-like image styles (Places365) or less consistent stylistic domains (DomainNet) leads to lower improvements, if not worse performance, than the baseline methods.


\section{Conclusions}

In this work, we introduce a new paradigm for Domain Generalization (DG) on three facets. We define two new measures, ICV and IDD, to quantitatively understand distributional shifts in stylistic domains based on Jensen-Shannon Divergence. We then present a novel precursor dataset SuperMarioDomains (SMD) sampled from scenes in video games, featuring more consistent categorical classes and dissimilar domains compared to ImageNet1K. We also demonstrate our new DG method SMOS that leverages the unique features of SMD as means to ground the training on DG benchmarks. We find that SMOS along with SMD reaches top performance on multiple DG benchmarks through significant improvements on abstract-styled target domains. SMOS has also been shown to qualitatively improve domain-invariant feature extraction by bringing distant domains within closer divergence in learned feature space. In the future, we would like to explore the application of our methodology on other tasks, such as Controllable Text-to-Image Generation or Visual Question Answering. 


\section*{Acknowledgements}
The authors acknowledge Research Computing at Arizona State University for providing HPC resources and support for this work.
This work was supported by NSF RI grants \#1750082 and \#2132724. 
The views and opinions of the authors expressed herein do not necessarily state or reflect those of the funding agencies and employers. 

{\small
\bibliographystyle{ieee_fullname}
\bibliography{references}
}

\end{document}